\documentclass{article}

\usepackage[final]{neurips_2024}

\usepackage{tabularray}

\usepackage[utf8]{inputenc} 
\usepackage[T1]{fontenc}    
\usepackage{hyperref}       
\usepackage{url}            
\usepackage{booktabs}       
\usepackage{amsfonts}       
\usepackage{amstext}
\usepackage{amsmath}
\usepackage{nicefrac}       
\usepackage{microtype}      
\usepackage{xcolor}         
\usepackage[pdftex]{graphicx}
\usepackage{multirow}

\title{Promoting cross-modal representations to improve multimodal foundation models for physiological signals}

\author{%
    Ching Fang \textsuperscript{1} \\
    Apple \\
    Columbia University \\
    \And
    Chris Sandino \\
    Apple \\
    \And 
    Behrooz Mahasseni \\
    Apple \\
    \And
    Juri Minxha \\
    Apple \\
    \And 
    Hadi Pouransari \\
    Apple \\
    \And
    Erdrin Azemi \\
    Apple \\
    \And
    Ali Moin \\
    Apple \\
    \And
    Ellen Zippi \textsuperscript{2} \\
    Apple \\
}

\begin{document}

\maketitle

\footnotetext[1]{Work completed during internship at Apple.} 
\footnotetext[2]{Corresponding author: ezippi@apple.com} 

\begin{abstract}
  Many healthcare applications are inherently multimodal and involve multiple types of physiological signals. As sensors for measuring these signals become more ubiquitous, it is increasingly important to improve machine learning methods that consume multimodal healthcare data. Pretraining foundation models is a promising avenue for success. However, methods for developing foundation models in healthcare are still early in exploration and it is unclear which pretraining strategies are most effective given the diverse set of physiological signals collected. This is in part due to challenges of multimodal learning with health data: data across many patients is difficult to obtain and expensive, and there is a lot of inter-subject variability. Furthermore, modalities are often heterogeneously informative across the downstream tasks of interest. Here, we explore these challenges in the PhysioNet 2018 Challenge dataset collected across 1,985 patients. We used a masked autoencoding objective to pretrain a multimodal model on the dataset. We show that the model learns representations that can be linearly probed for a diverse set of downstream tasks. We hypothesize that cross-modal reconstruction objectives are important for the success of multimodal training as they encourages the model to combine information across modalities. We demonstrate that adding modality drop in the input space improves model performance across downstream tasks. We also show that late-fusion models pretrained with contrastive learning objectives are not as effective as across multiple tasks. Finally, we analyze the representations developed in the model. We show how attention weights become more cross-modal and temporally aligned as a result of our chosen pretraining strategy. The learned embeddings also become more distributed in terms of the modalities that each unit in the model encodes. Taken together, our work demonstrates the utility of multimodal foundation models with health data, even across diverse physiological data sources. We further argue how more explicit means of inducing cross-modality may be valuable additions to any multimodal pretraining strategy.
\end{abstract}

\section{Introduction}
Healthcare applications often involve integrating information across many modalities. For instance, to diagnose sleep disorders, physicians may evaluate neural, muscular, and respiratory signals \citep{ibanez2018survey}. Adding to the complexity, the data used in healthcare spans a wide variety of formats (imaging data, time series, etc) and are collected from sensors placed on many different body locations \citep{acosta2022multimodal}. Many of these sensors for health data are becoming increasingly prevalent in everyday wearable devices \citep{cheol2018wearable, wu2019wearable, iqbal2021advances}. This technological advance is a promising opportunity for personalized healthcare and improving patient care. Thus, it is more and more important to leverage artificial intelligence to aid the interpretation of health data with heterogenous sensors.

In many settings, artificial intelligence has achieved unprecedented success in the development of multimodal foundation models \citep{jin2024efficient, bordes2024introduction, wadekar2024evolution}. For instance, models can now integrate information across language, vision, audio, and video to solve complex tasks and perform human-like feats of reasoning \citep{radford2021learning, alayrac2022flamingo, wu2023next, lu2024unified, mizrahi20244m}. Multimodal foundation models are pretrained in a self-supervised manner on vast amounts of data to link information across modalities. The representations developed by these models are useful for tasks that require multimodal understanding. After pretraining, these models may be further trained on a downstream task or the representations they produce can be used as is. Pretraining strategies often outperform models trained from scratch on the same tasks and require less labeled data \citep{jin2024efficient}. The success of multimodal foundation models in other domains suggests that similar advances can be achieved in healthcare settings.

There are further reasons to believe that health data in particular can benefit from foundation model strategies. Annotated data is limited in health data because clinical expertise is often necessary to create labels. Thus, the label efficiency of pretrained models is very useful in this setting. Furthermore, when considering wearable health devices, it becomes more important to develop models that are size-efficient. If a model pretrained on health data can successfully transfer its representations across many downstream tasks, this can greatly save on memory and runtime costs for wearable devices.

However, working with health data also introduces new types of challenges. Pretraining often consumes large amounts of unlabeled data, but patient privacy concerns limit the amount of large datasets available in this domain \citep{acosta2022multimodal, shaik2023survey}. In addition, the cost associated with deploying many health sensors can make large-scale data collection prohibitively expensive \citep{acosta2022multimodal}. Thus, it becomes less clear whether pretraining can be as effective as it is in settings like natural language, where large corpora are more widely available. In health applications, it is also common for certain modalities to vary greatly in their informativeness for different downstream tasks \citep{krones2024review}. This problem is exacerbated in wearable devices since different sensors may suffer from unequal amounts of noise, perhaps due to weaker contact or interference from other devices \citep{ates2022end, canali2022challenges}. This poses a challenge for developing general purpose models that can be used for diverse tasks.

Here, we investigate these challenges by pretraining a multimodal model in a publicly available dataset with 1,985 patients. We are specifically concerned with time series data collected from physiological signals measured overnight from patients. Our contributions are the following:
\begin{itemize}
    \item We explore the development of a multimodal foundation model in a dataset of diverse physiological signals: electroencephalography (EEG), electromyography (EMG), electrooculography (EOG), and electrocardiology (ECG). We demonstrate the strength of the learned representations in linear probe experiments on a disparate set of downstream tasks.
    \item We show how explicitly enforcing cross-modal reconstruction in the pretraining objective improves the quality of the learned representations over standard multimodal MAE. We also show how late-fusion models pretrained with contrastive learning does not effectively transfer across multiple tasks. 
    \item We analyze the learned representations to show that attention weights in the model become increasingly cross-modal under the pretraining objective we use. We also show that individual units in the model become more diversely tuned to the different modalities.
\end{itemize}

\begin{figure}
  \centering
  \includegraphics[width=\linewidth]{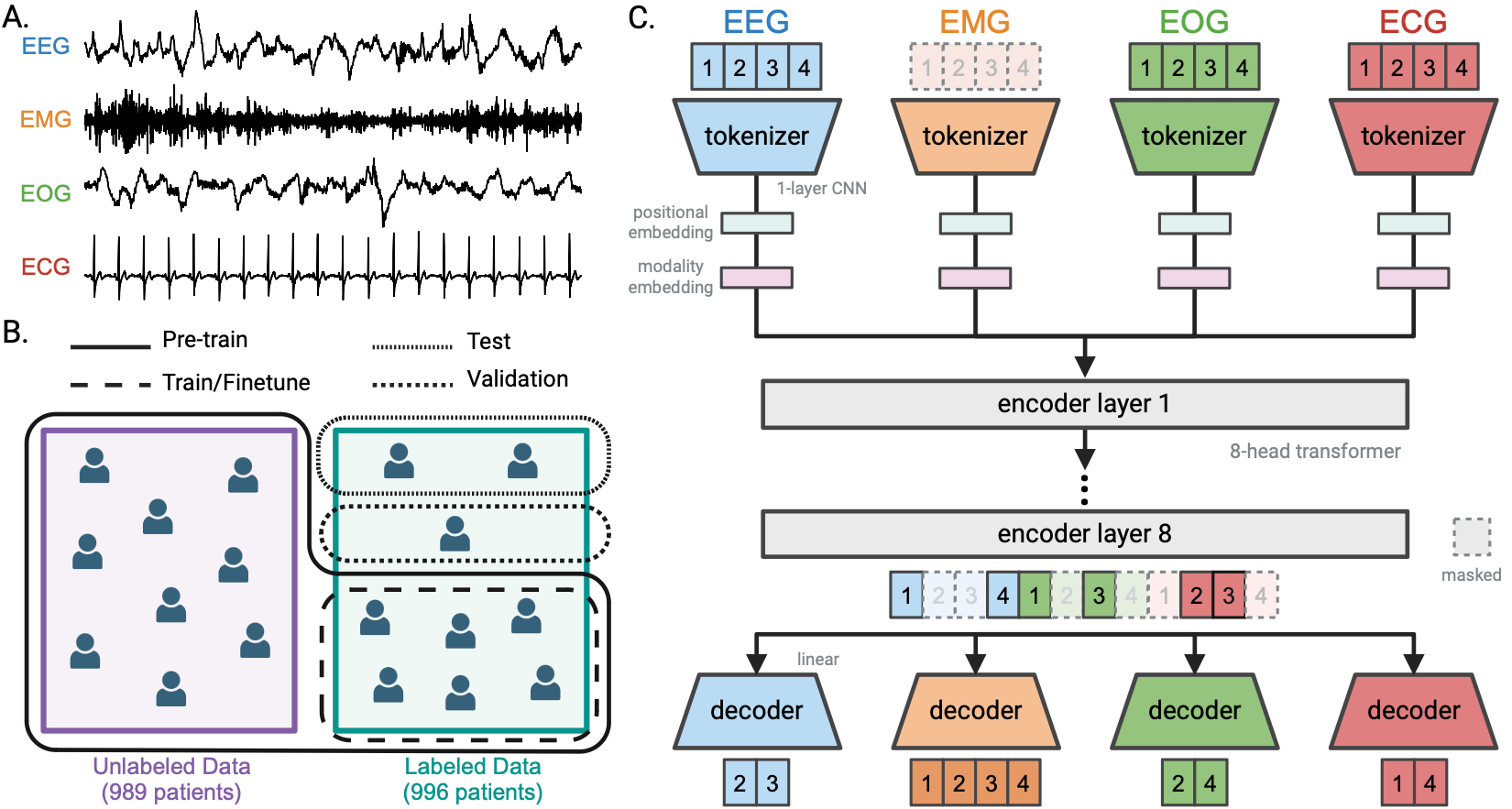}
  \caption{\textbf{A.} A 30-second sample from the training dataset. \textbf{B.} Data is split by patient identity for each part of the training procedure. The PhysioNet 2018 dataset consists of unlabeled data from 989 patients and labeled data from 996 patients, where each patient contributes 7.7 hours of data on average. The data for pretraining consists of all patients in the unlabeled dataset and 657 patients from the labeled dataset. The data for training and finetuning is drawn from the patients of the labeled dataset that were also used for pretraining. The data for the validation and test are drawn from the remaining patients of the labeled dataset not used for either pretraining or training. \textbf{C.} Diagram of the main pretraining strategy we use: multimodal masked autoencoding with modality drop in the input space. Tokenizers are modality-specific.}
\end{figure}

\section{Related Work}

Pretraining models with self-supervised objectives is a popular and effective strategy in machine learning \citep{ericsson2022self, gui2023survey}. After pretraining, the parameters of the model can be finetuned for some downstream task. Alternatively, another common approach is to freeze the pretrained model and train a lightweight readout head that uses the learned representations from the model to solve downstream tasks. This approach is especially attractive if efficiency in parameter tuning is a priority. In either case, a pretraining paradigm often outperforms training a model from scratch. Pretraining is especially useful if labeled data in the downstream task is limited as it provide a means for experimenters to define inductive biases on the model representations. Self-supervised strategies span several categories, including generative methods, contrastive learning, and autoencoding \citep{del2023applications, gui2023survey}. We limit our discussion to the latter two in the context of multimodal pretraining.  

Contrastive learning is a self-supervised learning framework where models are optimized such that representations of data in positive pairs become more similar while representations of data in negative pairs become more dissimilar \citep{chen2020improved, purushwalkam2020demystifying}. The definition of positive and negative pairs is crucial to the success of these methods. One way to define these pairs is to construct multiple ``views'' of a data sample through augmentations, like in the SimCLR algorithm \citep{chen2020improved, yuan2021multimodal}. Thus, a positive pair of data may be two different augmentations of a single data sample (negative pairs would then be constructed across different data samples). When working with multimodal data, another option is to consider each modality as a distinct view of a data sample. In this case, positive pairs can be constructed by comparing representations across modalities, as in CLIP-style pretraining \citep{radford2021learning, yuan2021multimodal, zhang2022contrastive}.

Masked autoencoding (MAE) is another popular pretraining strategy. In MAE, random patches of the input are masked, and the model must use the remaining portions of the input to reconstruct the masked portion \citep{he2022masked}. This method has been extended to settings with multimodal data, often to combine text data and vision data \citep{arici2021mlim, geng2022multimodal, bachmann2022multimae, zhao2023mamo, mizrahi20244m}. To do so, these models combine data across modalities early on so that representations are multimodally fused through layers of the model. The joint embeddings are then used for the MAE task to reconstruct inputs across all modalities. This structure inherently allows for the possibility of cross-modal reconstruction, as information from one modality can be used to reconstruct another. MAE methods can be more compute- and size- efficient due to the fused encoding structure used and the large amounts of data typically dropped out as a result of the masking strategy \citet{bachmann2022multimae, mizrahi20244m}. Of the above, the method used for our model is most similar to MultiMAE introduced in \citet{bachmann2022multimae}.

Both of these pretraining strategies have been applied to physiological signals, although examples are sparser than in other domains. We first discuss examples using contrastive learning strategies. \citet{abbaspourazad2023large} uses a large-scale Apple Watch dataset to classify demographics and health information from two modalities: photoplethysmography (PPG) and ECG. \citet{thapa2024sleepfm} used sleep data collected across EEG, EMG, ECG, and EOG sensors for downstream sleep-related classification tasks. \citet{raghu2022contrastive} uses cardiac and blood-related signals to predict mortality rate and pulmonary arterial pressure. Both \citet{abbaspourazad2023large} and \citet{raghu2022contrastive} use a SimCLR-like strategy through data augmentations, while \citet{thapa2024sleepfm} uses a CLIP-like strategy and construct data pairs across modalities.

In comparison to contrastive methods, MAE pretraining is less common for multimodal physiological signals. \citet{mathew2024foundation} uses MAE-pretraining in a model for phonocardiogram (PCG) and ECG data. The data was collected from digital stethoscopes, and the model was finetuned to classify signatures of cardiovascular disease. The closest example to our work is from \citet{liu2023frequency}, where a multimodal transformer model is pretrained on EEG, EMG, and EOG signals with a MultiMAE-like objective. However, this work was more limited in dataset size (100 patients in each pretraining dataset) and focused on one specific downstream task per pretrained model. In our work, we use a larger dataset with 1,985 patients and evaluate how well MultiMAE-pretrained models can perform on diverse downstream tasks. We later will make comparisons with contrastive methods as well.

A focus of our work is in encouraging cross-modal representation learning. This is inspired by works arguing that multimodal learning can be improved by optimizing for cross-modal reconstruction \citep{kleinman2023critical, hussen2020modality, hazarika2022analyzing}. While this objective is already present in the original MultiMAE algorithm, a simple way to further encourage cross-modal learning is to randomly drop modalities from the input \citep{hazarika2022analyzing, hussen2020modality, arici2021mlim, deldari2023latent}. This pressures the model to learn relationships across modalities in order to satisfy the reconstruction task. In the health data field, modality dropout strategies have been used to improve performance in tasks with missing modalities or heterogeneous noise, but they are still limited in their use in a general pretraining strategy. Furthermore, analyses of how representations are shaped by multimodal fusion are largely unexplored. We investigate both these questions in this work.

\section{Methods}
\subsection{Dataset}
We use the publicly available PhysioNet 2018 Challenge dataset \citep{ghassemi2018you}. This dataset consists of physiological signals collected during overnight sleep from 1,985 subjects. On average, each subject contributes 7.7 hours of recording \citep{ghassemi2018you}. The dataset contains many sensors, but here we focus on EEG, EMG, EOG, and ECG recordings (Figure 1A). For EEG, we use only the F3-M2 differential pair for our main results. We note that the signals from these sensors show distinct characteristics and are not obviously related (Figure 1A).

Patient demographics such as age and gender were also recorded in the dataset. The physiological data comprises of unlabeled data from 989 patients and labeled data from 996 patients. In the labeled set, 30-second contiguous windows were manually annotated by several certified sleep technologists into one of five sleep stages: wakefulness, stage 1, stage 2, stage 3, or rapid eye movement (REM). The same windows were also manually annotated for the presence of arousals (e.g. snores, vocalizations, respiratory effort, leg movement, etc.). Prior literature using this dataset mostly focus on the sleep staging task \citep{perslev2019u, banville2021uncovering, phan2021xsleepnet}, and comparisons to these works are discussed in the Appendix.

To prevent data leakage, data is split over patient identity. The pretraining dataset is comprised of of all 989 patients in the unlabeled set and 657 patients in the labeled set. The training/finetuning dataset for downstream tasks is comprised of the 657 patients in the labeled set that were used for pretraining (that is, the training/finetuning dataset is a subset of the pretraining dataset). The validation set and test set are constructed from the remaining 117 and 219 patients of the labeled set, respectively, and are not seen in either pretraining or training/finetuning. The validation set is used to select hyperparameters of the model and the test set is used for the evaluation scores reported in the results. A visualization of these data splits are in Figure 1B.

The signals are preprocessed with an anti-aliasing bandpass FIR filter then downsampled from 200 Hz to 100 Hz using decimation by 2. Specifically, EEG and EOG  signals were filtered to 0.1-30 Hz \citep{feng2021automatic, satapathy2024machine}. EMG and ECG signals were filtered to 0.1-70 Hz \citep{burns2007emg, feng2021automatic, satapathy2024machine}. All signals are then resampled to 100 Hz. We use 30-second samples of data for pretraining and for the downstream classification tasks. 

Three tasks are constructed from this dataset: (1) sleep scoring, (2) age classification, and (3) arousal identification. Sleep scoring is a $5$-way classification problem. Both arousal and age will be treated as a binary classification problem. In the age classification task, we aim to identify whether a patient's age is under 55 (the mean age) or not.

\begin{table}
  \caption{\textit{Balanced accuracy with linear probe evaluation: unimodal vs multimodal.} All models are pretrained before the encoder is frozen and representations are linearly probed for each task. We show the test balancy accuracy for a random guess (``Random''), for models trained entirely from scratch (``Scratch''), and for models pretrained and then linearly probed for the task (``Pretrained''). Note that "Pretrained-All' is a multimodal model pretrained with MultiMAE and input modality drop. Mean score and standard deviation for the three tasks are shown in columns. We additionally define an aggregate score which gives the average score over all tasks, normalized by the corresponding chance performance value (a score of 0 would indicate no improvement from chance). 500 patients are used in the training set for task finetuning. 5 random seeds are used in each training/finetuning stage. Asterisks indicate the best-performing unimodal model for each task.}
  \label{}
  \centering
\begin{tabular}{clcccc}
\multicolumn{2}{c}{ } & \multicolumn{1}{c}{Sleep} & \multicolumn{1}{c}{Age} & \multicolumn{1}{c}{Arousal} & \multicolumn{1}{c}{Aggregate} \\ 
\toprule
\multicolumn{1}{c}{Random} & \multicolumn{1}{c}{--} & 0.2 & 0.5 & 0.5 & 0.0 \\ 
\toprule
\multirow{5}{*}{Scratch} & EEG & 0.717 ± 0.003 & 0.641 ± 0.004 & 0.568 ± 0.093 & 1.0 ± 0.101 \\ 
& EMG & 0.461 ± 0.004 & 0.55 ± 0.006 & 0.538 ± 0.074 & 0.494 ± 0.076 \\ 
& EOG & 0.697 ± 0.006 & 0.626 ± 0.006 & 0.56 ± 0.082 & 0.952 ± 0.084 \\ 
& ECG & 0.279 ± 0.006 & 0.605 ± 0.022 & 0.516 ± 0.038 & 0.213 ± 0.025 \\ 
& All & 0.737 ± 0.003 & 0.626 ± 0.018 & 0.595 ± 0.013 & 1.042 ± 0.009 \\ 
\toprule
\multirow{5}{*}{Pretrained} & EEG & \textbf{0.745 ± 0.001}* & 0.662 ± 0.001 & 0.604 ± 0.093 & 1.085 ± 0.106 \\ 
& EMG & 0.442 ± 0.001 & 0.615 ± 0.003 & 0.533 ± 0.048 & 0.502 ± 0.052 \\ 
& EOG & 0.727 ± 0.001 & 0.653 ± 0.003 & 0.636 ± 0.071* & 1.07 ± 0.078 \\ 
& ECG & 0.339 ± 0.002 & 0.703 ± 0.002* & 0.526 ± 0.04 & 0.385 ± 0.042 \\ 
& All & 0.744 ± 0.001 & \textbf{0.719 ± 0.002} & \textbf{0.637 ± 0.081} & \textbf{1.144 ± 0.09} \\ 
\bottomrule
\end{tabular}

\end{table}

\subsection{Model architecture}
Our model architecture is based off that of the vision transformer \citep{alexey2020image}. Modality-specific tokenizer layers are followed by fused encoding layers, so that multimodal information is fused early on (Figure 1C). The input to the model is a 30 second time series from multiple sensors sampled at 100 Hz. We divide each time series into 30 chunks that are one second each. These chunks are then fed to the tokenizer layers. Tokenizers are trained for each modality and consist of one convolutional layer and one linear layer. Specifically, each signal chunk first passes through a 1D convolutional layer (with 64 channels and kernel size of 21) before a max pooling operation. Then, a linear layer projects each token into a 512-dimensional embedding space. This is followed by layer normalization to ensure signals from all modalities have comparable scales. Given 1,985 patients with an average of 7.7 hours of recording time each, the total dataset size is 1,834,140.

To summarize, the output of a tokenizer for one modality is 30 tokens with embedding dimension $D=512$. Sinusoidal positional embeddings and a learnable modality embedding are then added to each token. Finally, tokens across modalities are fused through concatenation.

This fused vector is then passed to the joint encoding layers, which is comprised of eight transformer layers with multi-head self-attention \citep{vaswani2017attention} and normalization before attention layers \citep{xiong2020layer}. Each transformer layer has 8 heads, and each layer has a 10\% dropout rate during training over attention weights and projection weights.

\subsection{Pretraining objectives}
We use a multimodal masked autoencoding (MAE) objective similar to MultiMAE from \citet{bachmann2022multimae}. As mentioned above, tokens across all modality tokenizers are fused via concatenation. In MultiMAE, a fixed portion of these tokens are masked at uniform and dropped from the fused vector. We use a 70\% masking rate (see Appendix for how the mask rate was selected). The remaining unmasked tokens are passed into the encoder and processed. To prepare the input for the decoder layers, the tokens that are output from the encoder are then interleaved with learnable mask tokens. Values in the mask token are initialized from $\mathcal{N}(0, 0.02)$ with truncation at $[-2,2]$. These learnable mask tokens act as placeholders for the signal to be reconstructed (i.e., the dropped tokens). Mask tokens are inserted in the location of the previously dropped tokens. Positional information is preserved by adding the appropriate positional embedding to the newly interleaved mask tokens.

A decoder is trained for each modality to reconstruct the original signal. Each decoder consists of a cross-attention layer and a transformer layer before a linear projection. The input into each modality decoder is the subset of tokens from the encoder output that corresponds to that modality. Cross-modal reconstruction is enabled through the cross-attention layer, where the input is the query and the entire encoder output is passed as keys/values. The linear layer projects each token from the embedding dimension (512) to the original signal dimension (100). The loss is calculated only over the reconstructed signal corresponding to the dropped tokens.

To encourage additional cross-modal interactions, we also use input modality drop during pretraining (Figure 1B) \citep{hazarika2022analyzing, hussen2020modality, arici2021mlim, deldari2023latent}. In every batch, one randomly chosen modality is completely dropped on top of the typical MultiMAE uniform masking over tokens.

In later experiments we will make comparisons with contrastive learning objectives, resulting in modifications to the pretraining loss and the model architecture. In this case, the model will be converted to a late fusion structure that is typical for models trained with contrastive objectives. Further details can be found in the corresponding results section (\S 4.3) and Appendix F. 

\subsection{Finetuning}
We are most interested in understanding how well representations learned by the pretrained model can transfer to multiple tasks. As such, after pretraining, we discard the decoder and freeze the encoder. The output of the encoder is layer normalized and average pooled over the token dimension. This $512$-dimensional vector is then passed to a linear classification head. A classification head is trained for each of the downstream tasks with weighted cross entropy loss to account for class imbalance. This is most relevant for the arousal detection task, where arousal events are extremely rare (2.7\% of data samples). Although not the focus of this paper, we also conduct full finetuning experiments where both encoder and classifier parameters are trained (Appendix E).

\subsection{Optimization}
Models are pretrained for 2000 epochs or until a fixed compute time of 10 days is exceeded. For finetuning, models are trained for 200 epochs. Learning rates were scheduled with 10 epochs of linear warmup to $1\times10^{-4}$ and cosine annealing thereafter \citep{loshchilov2016sgdr}. We used the AdamW optimizer \citep{loshchilov2017decoupled}. The model checkpoint chosen for evaluation was from the epoch where the lowest validation error was achieved, except in the case of pretraining the MultiMAE model with input modality drop. In this case, the validation error was quite noisy and the most recent checkpoint was chosen instead. Additional details can be found in the Appendix.

\section{Experiments}

\begin{table}
    \caption{\textit{Linear probe evaluation on all three tasks, comparing multimodal pretraining strategies}. All models are pretrained before the encoder is frozen and representations are linearly probed for each task. Mean test score and standard deviation for the three tasks are shown in columns. Aggregate score is defined as in Table 1. 500 patients are used in the training set for task finetuning. 5 random seeds are used in each training/finetuning stage.}
    \label{}
    \centering
\resizebox{\linewidth}{!}{%
    \begin{tabular}{cccccc}
    \toprule
    \multirow{2}{*}{Pretraining Strategy} & \multicolumn{2}{c}{Sleep} & & \multicolumn{2}{c}{Age} \\
    \cline{2-3} \cline{5-6} 
    & Balanced Acc. & Cohen Kappa & & Balanced Acc. & AUROC \\
    \toprule
    Contrastive CLIP-style (LOO) & 0.708 ± 0.0004 & 0.572 ± 0.001 & & 0.643 ± 0.004 & 0.705 ± 0.006 \\
    Contrastive CLIP-style (Pairwise) & 0.703 ± 0.001 & 0.559 ± 0.001 & & 0.646 ± 0.0003 & 0.698 ± 0.0003 \\
    Contrastive SimCLR-style & 0.656 ± 0.001 & 0.52 ± 0.001 & & 0.624 ± 0.009 & 0.673 ± 0.015 \\
    MultiMAE Only & 0.734 ± 0.001 & 0.618 ± 0.001 & & 0.684 ± 0.001 & 0.758 ± 0.001 \\
    MultiMAE + Input Mod. Drop & \textbf{0.744 ± 0.001} & \textbf{0.63 ± 0.002} & & \textbf{0.719 ± 0.002} & \textbf{0.785 ± 0.002} \\
    \bottomrule
    \end{tabular}}

    \vspace{1em}
\resizebox{\linewidth}{!}{%
    \begin{tabular}{ccccccccc}
    \toprule
    \multirow{2}{*}{Pretraining Strategy} & \multicolumn{2}{c}{Arousal} & & & & \multirow{2}{*}{Aggregate Score} & & \\
    \cline{2-3}
    & Balanced Acc. & AUROC & & & & & &\\
    \toprule
    Contrastive CLIP-style (LOO) & \textbf{0.71 ± 0.002} & \textbf{0.776 ± 0.001} & & & & 1.082 ± 0.005 & & \\
    Contrastive CLIP-style (Pairwise) & 0.708 ± 0.002 & 0.772 ± 0.001 & & & & 1.075 ± 0.002 & &\\
    Contrastive SimCLR-style & 0.585 ± 0.048 & 0.616 ± 0.070 & & & & 0.900 ± 0.040 & & \\
    MultiMAE Only & 0.604 ± 0.089 & 0.638 ± 0.136 & & & & 1.082 ± 0.062 & & \\
    MultiMAE + Input Mod. Drop & 0.637 ± 0.081 & 0.677 ± 0.128 & & & & \textbf{1.144 ± 0.058} & & \\
    \bottomrule
    \end{tabular}}
    
\end{table}

\subsection{A pretrained multimodal model develops representations that support a diverse set of tasks in the PhysioNet18 dataset.}

We first assess the extent to which pretraining and multimodal learning benefits downstream task performance in this dataset. We evaluate performance on the three tasks when both unimodal and multimodal models are trained from scratch. The balanced accuracy achieved by these models on the test set is shown in Table 1 (``Scratch'' rows). In addition to the three tasks, we also define an aggregate score to highlight models that perform well across all tasks. The aggregate score is defined as $\frac{1}{N}\sum_i^N \frac{s_{i}-r_{i}}{r_{i}}$, where $s_i$ is the average test score on task $i$, $r_i$ is the chance level performance for task $i$, and $N=3$ is the total number of tasks. Scores are measured using balanced accuracy. We see that the multimodal model performs overall better than any of the unimodal models (compare aggregate scores), although its performance on the sleep classification task slightly lags behind the unimodal EEG model. 

We next examine the benefits of pretraining the model and transferring the learned representations to each of the downstream tasks. We first pretrain the unimodal models with masked autoencoding. The test scores for these models are shown in Table 1 as well (``Pretrained'' rows). Pretraining seems to benefit all models, whether unimodal or multimodal. Interestingly, the pretrained unimodal models reveal that a different modality is most informative for each task: EEG is more effective for sleep staging, ECG for age classification, and EOG for arousal classification. 

We then pretrain a multimodal model with MultiMAE and input modality drop. We evaluate this model on the downstream tasks (``Pretrained, All'' in Table 1). The multimodal model outperforms the unimodal model in age classification and arousal classification, and performs very similarly to EEG in the sleep staging task (Table 1). We find that the multimodal model performs well in all tasks and achieves a higher aggregate score, despite the imbalance in modality dominance across tasks.

Notably, the improvement in aggregate score obtained by the multimodal model is greater when training data is more limited (Appendix D). Given full-finetuning, though, the differences across models are more minimal (Appendix E). 

\subsection{Adding input modality drop to MAE pretraining is important for downstream task performance.}

We chose our particular pretraining strategy with the hypothesis that encouraging multimodal fusion improves performance in the downstream tasks. We investigate whether this is the case by first testing the importance of using input modality drop (which theoretically should result in more cross-modal learning). We compare task performance to that of a standard MultiMAE strategy, which does not include input modality drop (Table 2). We see that removing input modality drop causes a performance drop in downstream tasks (compare ``MAE Only'' to ``MAE + Input Mod. Drop'' in Table 2). In fact, without input modality drop, MultiMAE underperforms the most informative unimodal models across all tasks. Overall, dropping modalities in the input appears to be a simple and effective means to increase performance over standard MultiMAE.

\begin{figure}
  \centering
  \includegraphics[width=\linewidth]{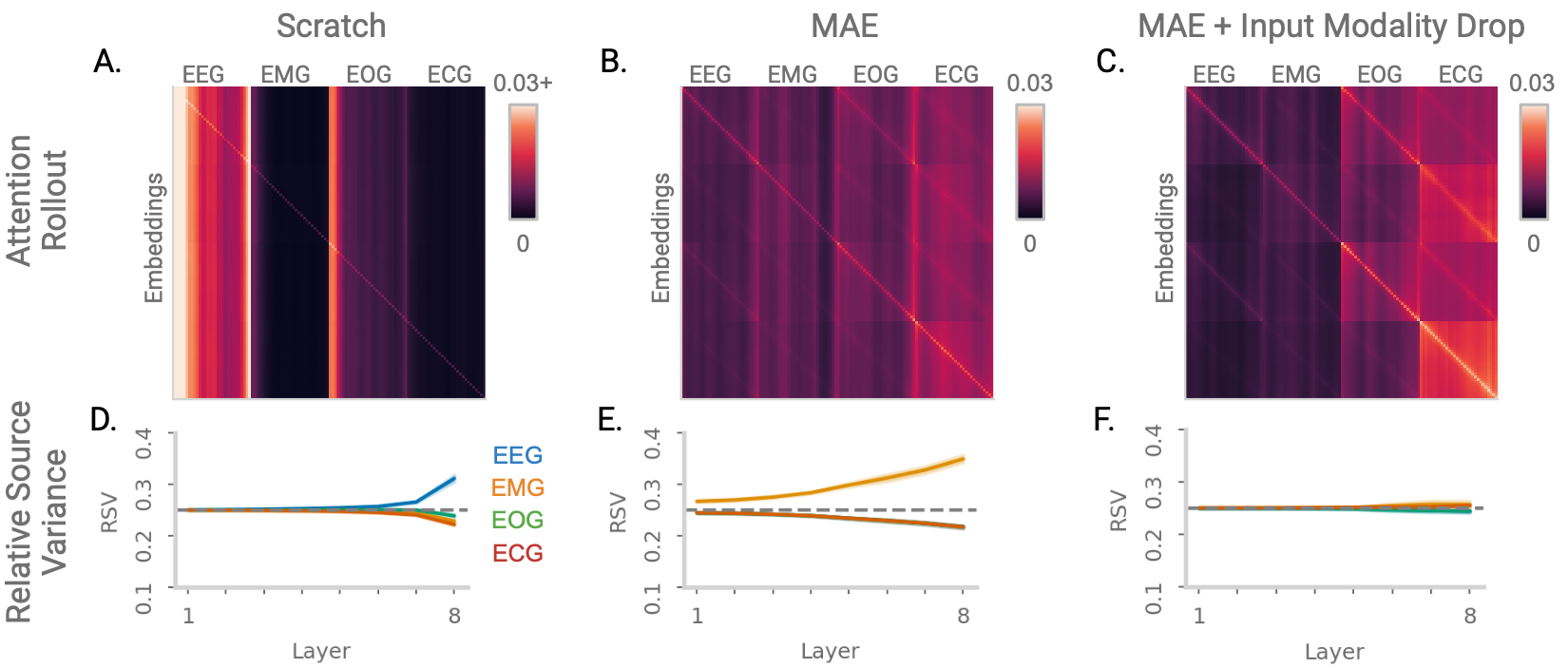}
  \caption{Measures of modality fusion across model representations. \textbf{A.} Attention rollout from tokens in the embeddings to tokens in the input. Here, the model is trained from scratch on sleep staging. Values are capped at 0.03 for comparisons with (BC). \textbf{B.} As in (A), but for the model pretrained with MAE. \textbf{C.} As in (A), but for the model pretrained with MAE and input modality drop. \textbf{D.} Relative source variance (RSV) of units across layers of the model in (A) to each of the four modalities. 95\% confidence intervals shown, over 512 units in each embedding vector. \textbf{EF.} As in (D), but for the models in (B) and (C), respectively.}
\end{figure}

\subsection{Late fusion models with contrastive learning objectives are more variable in performance.}
Multimodal fusion is additionally encouraged in the MultiMAE model through the early fusion architecture, where representations across modalities are mixed early in the network. We next make comparisons to models with a late fusion structure where representations across modalities are not mixed except in the decoders for downstream tasks (Appendix F). To do so, we pretrain late fusion models with contrastive learning, a common pretraining objective for these types of model. 

We first test SimCLR-style multiview contrastive learning \citep{chen2020improved, purushwalkam2020demystifying}, with particular inspiration from \citet{raghu2022contrastive}. We randomly generate augmentations for all input data samples (using the same signal augmentations from \citet{raghu2022contrastive}). Positive pairs are defined as representations from adjacent time windows. We find that the SimCLR-style model underperform standard MultiMAE in all tasks (Table 2). This may indicate that defining desired relationships between of modality embeddings is important for the performance of a contrastive learning model.

We next test CLIP-style pretraining to assess the benefits of using modality contrast in the contrastive learning loss \citep{radford2021learning, yuan2021multimodal, zhang2022contrastive}. We will use two objectives defined in \citet{thapa2024sleepfm}, a previous work in physiological signals that inspired our approach here. \citet{thapa2024sleepfm} defined a pairwise loss and a leave-one-out (LOO) loss:
\setlength{\abovedisplayskip}{6pt}
$$l^{pair}_{ijk}=-\log \frac{\exp(\text{sim}(x^i_k, x^j_k))*\tau}{\sum^N_{m=1}\exp(\text{sim}(x^i_k, x^j_m))*\tau}
\ \ \ \ \ \ \ \ \ \ \ \ \ 
l^{LOO}_{ik}=-\log \frac{\exp(\text{sim}(x^i_k, \bar{x}^{\neq i}_k))*\tau}{\sum^N_{m=1}\exp(\text{sim}(x^i_k, \bar{x}^{\neq i}_m))*\tau}$$
for modalities $i$ and $j$, sample $k$, temperature $\tau$, and modality embedding $x$. $N$ is the total number of samples, and $\bar{x}^{\neq i}_k$ is the average of representations that are not modality $i$ given data sample $k$. We find that both CLIP-style models underperform even standard MultiMAE in sleep and age classification (Table 2). Surprisingly, the contrastive model does extremely well in arousal classification. However, in terms of aggregate performance, using MultiMAE with input modality drop is still most effective out of the strategies we tested.

Despite these results, we speculate that developing new formulations of contrastive learning may improve task performance. These methods are highly sensitive to the choice of positive and negative pairs. It may be that contrastive methods in multimodal biosignals require domain-specific design to reach their full potential.

\subsection{MAE + input modality drop encourages cross-modal fusion in attention weights and model representations.}

Finally, we wanted to understand whether our intuition about cross-modal fusion was indeed reflected in the representations developed by the model. We first examine the attention weights of the model to understand how much each output token from the encoder is influenced by input tokens from each modality. We use a method called attention rollout \citep{abnar2020quantifying}. Attention rollout accounts for the effects of the residual layers by defining the attention at layer $l$ as a sum of the raw attention weights and the identity matrix: $A_{l}=0.5 W_{l} + 0.5 I$ where $W_{l}=\text{softmax}(Q_{l} K_{l}^{T})$. Thus, to obtain the attention of the output embedding to the inputs, attention weights are rolled out across model layers: $A_L * A_{L-1} * \dots * A_2 * A_1$, for $L$ layers in the encoder.

We plot the results of attention rollout first for a multimodal model trained from scratch on sleep-scoring (Figure 2A). The attention matrix develops strong vertical bands, indicating that model embeddings attend to specific tokens without any context-specificity. In this case, EEG and EOG tokens are most dominant. We next plot the attention matrix for a model pretrained with MultiMAE and MultiMAE with input modality drop (Figure 2BC). The attention weights are more evenly spread across the matrix, indicating greater cross-modal attention, although some sparse vertical bands can still be observed. This can be interpreted as greater context-specificity in the attention weights. We also observe an additional effect from both MultiMAE models where attention weights become more temporally aligned. That is, tokens largely attend to other tokens that occurred around the same window of time (Figure 2BC), an effect that is also visible when examining the raw attention matrices $W_l$ (Appendix).

Although attention rollout allowed us to better understand the benefits of MultiMAE pretraining, it is unclear how input modality drop affects representations. To further investigate this, we next analyze individual embedding units in the model to see how tuned they are to different modalities. We use relative source variance (RSV), which quantifies the variance in the activity of a unit due to a particular input modality  \citep{kleinman2023critical}. As an example, assume we want to calculate RSV due to EEG. First, let $x_{EEG} \sim X_{EEG}$ be a sample of EEG data from the dataset (with similar notation for all other modalities). The source variance of a unit $a$ due to EEG when all other modalities $j$ are fixed at samples $x_j$ is defined as 
\setlength{\abovedisplayskip}{0.5em}
\setlength{\belowdisplayskip}{0.75em}
\begin{multline*}
        SV_a (X_{EEG}, x_{EMG}, x_{EOG}, x_{ECG}) = \\
        \text{Var}(f(X_{EEG}, X_{EMG}=x_{EMG}, X_{EOG}=x_{EOG}, X_{ECG}=x_{ECG})_a)
\end{multline*}
where $f$ gives the output embedding from the encoder, averaged over tokens. Symmetrically, source variance can also be defined for the other modalities. Taking the softmax over these source variances for a unit $a$ gives the relative source variance of $a$. Thus if unit $a$ is uniformly tuned to all input sources, it would have a RSV value of 0.25 for each modality.

We first measure the RSV values of embedding units in the model trained from scratch on the sleep staging task (Figure 2D). We find that representations become more tuned for EEG in the later layers of the model. This is likely because EEG is more informative for the task and thus the decoder places greater emphasis on EEG over the other modalities. We next measure the RSV values for the MAE-pretrained model (Figure 2E). Interestingly, we see that across layers, units in the model become increasingly tuned to EMG input. This is likely because the model struggles most to reconstruct EMG (Appendix) and thus places greater representation weight onto that modality. In contrast, the model trained with MAE and modality drop is equally tuned to all modalities across all layers (Figure 2F).

\section{Limitations and Discussion}
We have shown the strength of a foundation model-style approach using physiological data with a diverse set of downstream tasks. We compare a variety of approaches and argue that explicitly incorporating objectives that promote cross-modal reconstruction greatly improves representation quality for solving downstream tasks. Specifically, we find that incorporating input modality drop is a simple, yet especially effective strategy. We note that making comparisons with other datasets would be additionally informative, especially since multimodal fusion strategies are often dependent on the dataset and task at hand \citep{ma2022multimodal}. In addition, developing a large range of downstream tasks will provide better insights into the strengths of different pretraining strategies and help identify those that are especially useful for general purpose training. 

\bibliography{references}
\newpage

\appendix

\Large \textbf{Appendix} \normalsize

\section{Additional training details}
Code was written with PyTorch Lightning. All model training was executed on NVIDIA Tesla V100S GPUs with 32 GB of memory. In the multimodal model, we used 8 GPUs for pretraining, 8 GPUs for full finetuning, and 4 GPUs for linear finetuning. For additional efficiency, we use Lightning's automatic mixed-precision training. Pretraining to 2000 epochs took around 7-10 days (depending on the exact pretraining strategy). To run 200 epochs of full finetuning and linear finetuning on 500 patients in the training set, it took around 6 hours and 4 hours, respectively.

\section{Comparison to prior baselines in the PhysioNet18 dataset}
How does our model fare compare to other works that have used this dataset? An exact comparison is difficult since prior works with the PhysioNet18 dataset use different splits and different combinations of modalities or channels. However, we discuss a few examples here, all of which are concerned with the sleep staging task.

\citet{banville2021uncovering} use the F3-M2 and F4-M1 channels of EEG in a model pretrained with contrastive learning. The authors were also interested in limited training data. They found that, with 595 patients in the training data, the balanced accuracy achieved by their model on their test set was 72.3\%. 
\citet{phan2021xsleepnet} use one channel each of EEG, EOG, and EMG. They train bidirectional RNNs on 944 patients and report the 5-fold cross validated score as a Cohen's Kappa score of 0.847. \citet{perslev2019u} also use 944 patients and report the 5-fold cross validated score as a F1 score of 0.77. We note that, in our hands, MultiMAE + input modality drop trained on 500 patients achieve a F1 score of 0.72 on our test set. Further examples from other works can be found in \citep{phan2022automatic}.

\section{Selecting pretraining hyperparameters}
\begin{figure}[h]
  \centering  \includegraphics[width=\linewidth]{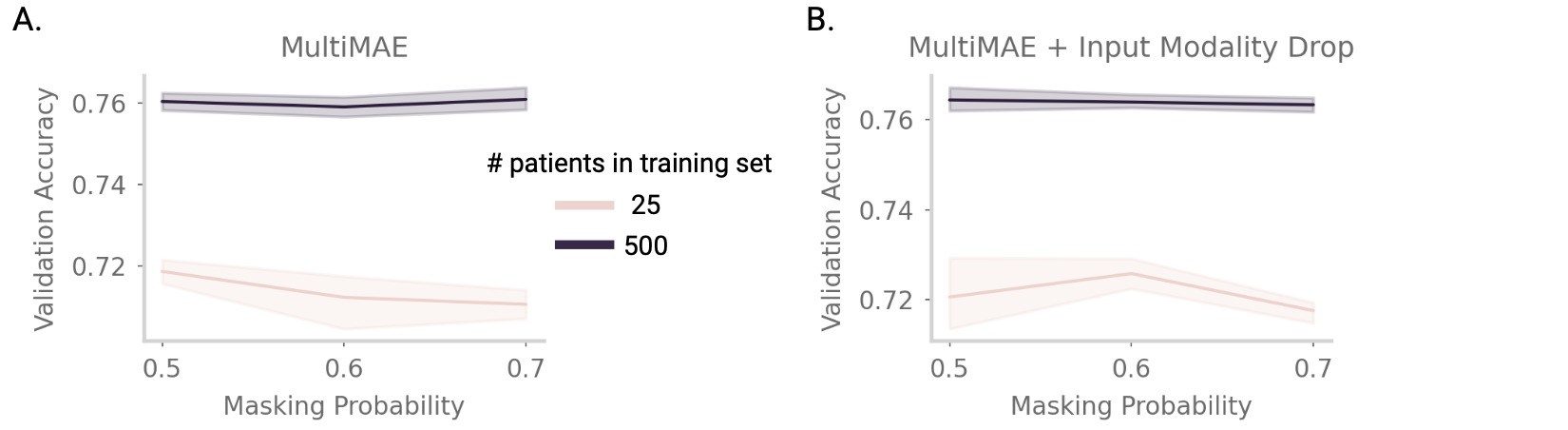}
  \caption{Hyperparameter selection in MAE models. \textbf{A.} Validation set accuracy score in the sleep staging task, with full-finetuning. Here, we show models pretrained with only MultiMAE-only. The x-axis shows the masking probability. \textbf{B.} As in (A), but for the model pretrained with MultiMAE and input modality drop.}
\end{figure}

We show validation scores of the MultiMAE model (Figure 3A) and the MultiMAE + input modality drop model (Figure 3B).

We begin with selecting the masking ratio for the model with input modality drop. If the masking ratio is $p$ the overall masking ratio is $\frac{1}{N} + \frac{N-1}{N} p$, where $N$ is the number of modalities. This expression arises from the modality dropping that occurs at each batch. By examining Figure 3B, it appears that using a masking probability of 0.6 is best, although this difference in performance across masking ratios is only visible in the low data regime (i.e., 25 patients in the training set). Thus, the overall masking ratio is 0.7.

Thus we select a masking ratio of 70\% for the MultiMAE model (Figure 3A) as it allows for clear comparison with MultiMAE + input modality drop. We also note that, with 500 patients in the training set, the choice of masking probability does not clearly affect the downstream task performance of the MultiMAE model. Thus, both models have the same amount of tokens masked during pretraining, with the only difference due to the distribution of masking across tokens.

For visualization of the pretraining performance, we show example reconstructions made by the MultiMAE model on two samples from the training set (Fig 4). The model clearly struggles the most with reconstructing the EMG signal (this is also reflected in the mean squared error values, although those are not shown here).

\begin{figure}[h]
  \centering  \includegraphics[width=\linewidth]{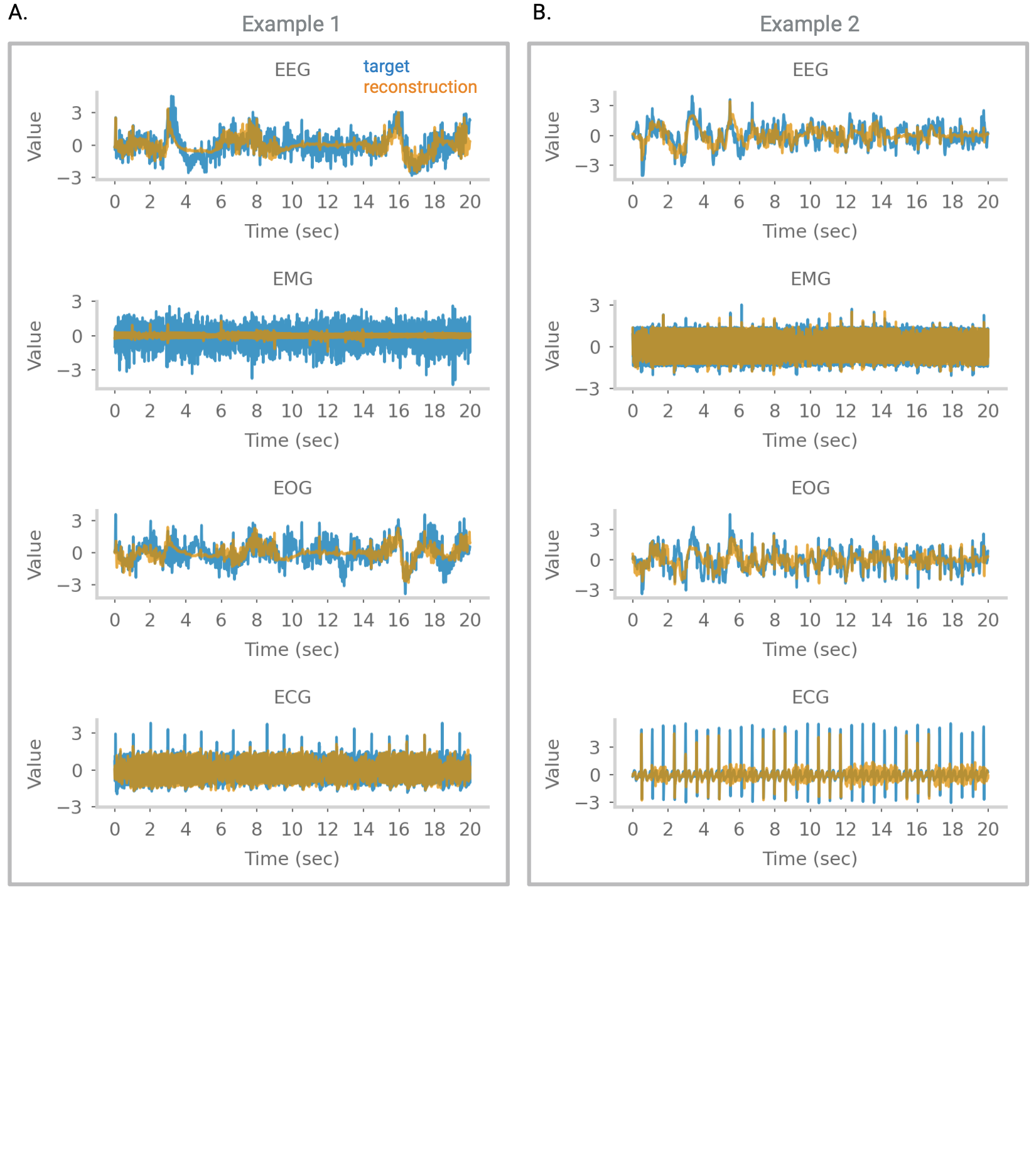}
  \caption{Reconstruction performance of MultiMAE model with 70\% masking. \textbf{A.} A random sample from the training data, with target signals in blue and reconstructed signals in orange. Plot is truncated at 20 seconds for visualization purposes. \textbf{B.} As in (A), but for another random sample}
\end{figure}

\newpage

\section{Additional results: limited training data during finetuning}

\begin{table}[h]
  \caption{\textit{Unimodal vs multimodal performance, with limited training data.} As in Table 1, but with 25 patients in the training data set for each of the three classification tasks.}
  \label{}
  \centering
\begin{tabular}{clcccc}
\multicolumn{2}{c}{ } & \multicolumn{1}{c}{Sleep} & \multicolumn{1}{c}{Age} & \multicolumn{1}{c}{Arousal} & \multicolumn{1}{c}{Aggregate} \\ 
\toprule
\multicolumn{1}{c}{Random} & \multicolumn{1}{c}{--} & 0.2 & 0.5 & 0.5 & 0.0 \\ 
\toprule
\multirow{5}{*}{Pretrained} & EEG & 0.637 ± 0.039 & \textbf{0.616 ± 0.007} & 0.52 ± 0.077 & 0.819 ± 0.096 \\ 
& EMG & 0.332 ± 0.012 & 0.524 ± 0.026 & 0.512 ± 0.017 & 0.244 ± 0.014 \\ 
& EOG & 0.635 ± 0.004 & 0.605 ± 0.005 & 0.595 ± 0.05 & 0.859 ± 0.049 \\ 
& ECG & 0.255 ± 0.006 & 0.545 ± 0.021 & 0.511 ± 0.022 & 0.129 ± 0.008 \\ 
& All & \textbf{0.688 ± 0.002} & 0.571 ± 0.004 & \textbf{0.597 ± 0.062} & \textbf{0.925 ± 0.067} \\ 
\bottomrule
\end{tabular}
\end{table}

\section{Additional results: full finetuning}

\begin{table}[h]
  \caption{\textit{Unimodal vs multimodal performance, with full-finetuning.} As in Table 1, but parameters of the encoder are also finetuned along with training of the classification head.}
  \label{}
  \centering
\begin{tabular}{clcccc}
\multicolumn{2}{c}{ } & \multicolumn{1}{c}{Sleep} & \multicolumn{1}{c}{Age} & \multicolumn{1}{c}{Arousal} & \multicolumn{1}{c}{Aggregate} \\ 
\toprule
\multicolumn{1}{c}{Random} & \multicolumn{1}{c}{--} & 0.2 & 0.5 & 0.5 & 0.0 \\ 
\toprule
\multirow{5}{*}{Pretrained} & EEG & \textbf{0.747 ± 0.003} & 0.656 ± 0.01 & 0.58 ± 0.02 & 1.069 ± 0.013 \\ 
& EMG & 0.457 ± 0.009 & 0.618 ± 0.006 & 0.562 ± 0.006 & 0.549 ± 0.018 \\ 
& EOG & 0.733 ± 0.001 & 0.637 ± 0.005 & 0.581 ± 0.011 & 1.033 ± 0.009 \\ 
& ECG & 0.341 ± 0.01 & 0.67 ± 0.018 & 0.55 ± 0.014 & 0.382 ± 0.01 \\ 
& All & 0.746 ± 0.005 & \textbf{0.694 ± 0.009} & \textbf{0.588 ± 0.023} & \textbf{1.098 ± 0.015} \\ 
\bottomrule
\end{tabular}
\end{table}

\pagebreak

\section{Contrastive Pretraining}
Here, we give more details of the contrastive pretraining strategy.

\begin{figure}[h]
  \centering
  \includegraphics[width=\linewidth]{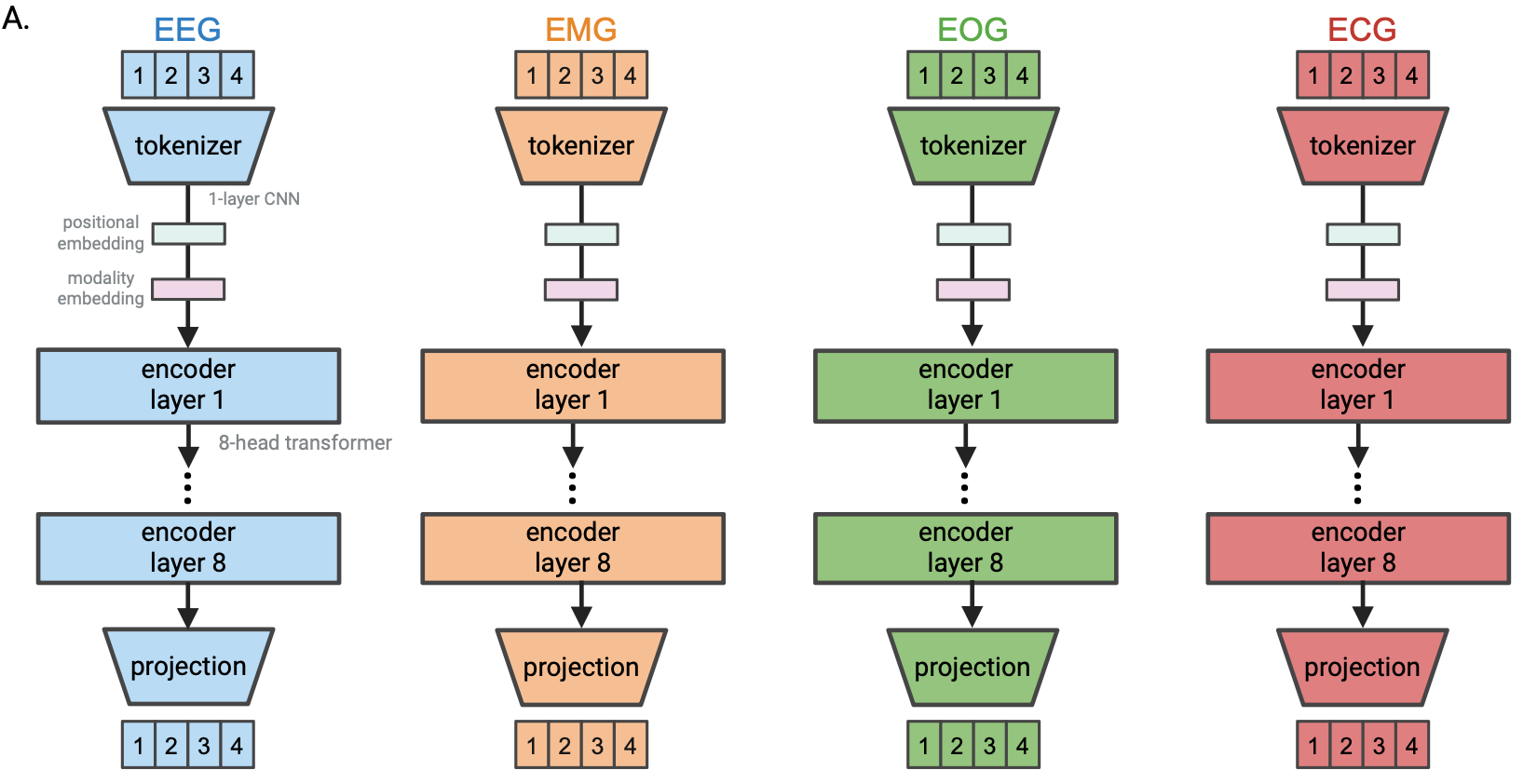}
  \caption{Contrastive learning architecture.}
\end{figure}

\subsection{SimCLR-style}
We use the loss functions introduced in \citet{raghu2022contrastive}. We note a few differences from our implementations and that of \citet{raghu2022contrastive}. One is that we use a transformer architecture to allow for comparisons with the MultiMAE models we tested. We also do not have structured data or static features. We use the same fixed temperature as in \citet{raghu2022contrastive}. Furthermore, we also add a MLP projection head before the embeddings are passed to the contrastive loss. As before, the encoder outputs are 512-dimensional. The projection head consists of a hidden layer of dimension 256 before projection into 128 dimensions. For a given data sample, the representation we use for contrastive learning is the concatenated representations across the four modalities. Specifically, the representation is the concatenation of the output of the four projection layers. These are the representations used in the similarity calculations. Finally, As in \citet{raghu2022contrastive}, the projection head is discarded after pretraining.

\subsection{CLIP-style}
We use the loss functions introduced in the SleepFM paper of \citet{thapa2024sleepfm}. The architecture we use is the same as in the SimCLR-style models. Besides the difference of our transformer architecture, another difference between our implementation and that of \citet{thapa2024sleepfm} is that we use (as in the SimCLR-like model) a fixed temperature parameter and MLP projection head. We found the use of a fixed temperature and projection head led to better validation set performance in the downstream tasks, which is why we introduce these extra details.

\newpage

\section{Raw attention matrices}
Raw attention matrices of the models shown in Figure 2A-C. These matrices correspond to $W_l$ in the expression for attention rollut given in \S 4.4 The matrix for each layer is averaged over the 8 heads.

\begin{figure}[h]
  \centering
  \includegraphics[width=\linewidth]{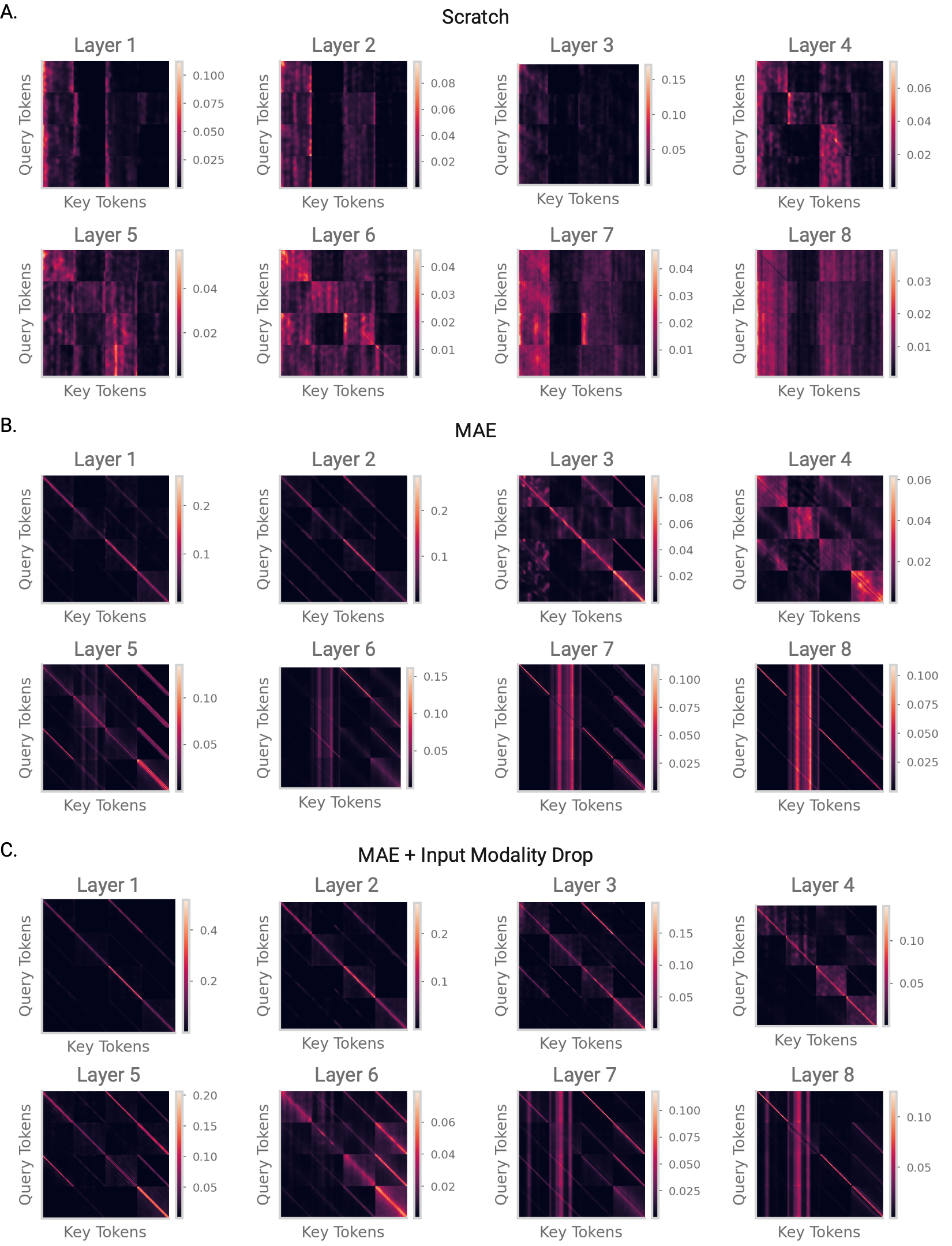}
  \caption{Raw attention matrices.}
\end{figure}

\section{Additional Evaluation Scores}
Same as in Table 2, but with additional metrics.

\begin{table}
  \caption{Sleep}
  \label{}
  \centering
  \begin{tabular}{lccc}
    \toprule
    Pretraining Strategy & Balanced Acc. & Cohen Kappa & F1 \\ 
    \toprule
    Contrastive CLIP-style (LOO) & 0.708 ± 0.0 & 0.572 ± 0.001 & 0.67 ± 0.001 \\ 
    Contrastive CLIP-style (Pairwise) & 0.703 ± 0.001 & 0.559 ± 0.001 & 0.658 ± 0.001 \\ 
    Contrastive SimCLR-style & 0.656 ± 0.001 & 0.52 ± 0.001 & 0.632 ± 0.001 \\ 
    MultiMAE + Modality Drop & \textbf{0.744 ± 0.001} & \textbf{0.63 ± 0.002} & \textbf{0.718 ± 0.002} \\
    \bottomrule
  \end{tabular}
\end{table}

\begin{table}
  \caption{Age}
  \label{}
  \centering
  \begin{tabular}{lccc}
  \toprule
    Pretraining Strategy & Balanced Acc. & AUROC & F1 \\
    \toprule
    Contrastive CLIP-style (LOO) & 0.643 ± 0.004 & 0.705 ± 0.006 & 0.655 ± 0.004 \\ 
    Contrastive CLIP-style (Pairwise) & 0.646 ± 0.0 & 0.698 ± 0.0 & 0.655 ± 0.0 \\ 
    Contrastive SimCLR-style & 0.624 ± 0.009 & 0.673 ± 0.015 & 0.635 ± 0.009 \\ 
    MultiMAE & 0.684 ± 0.001 & 0.758 ± 0.001 & 0.694 ± 0.001 \\ 
    MultiMAE + Modality Drop & \textbf{0.719 ± 0.002} & \textbf{0.785 ± 0.002} & \textbf{0.728 ± 0.001} \\
    \bottomrule
  \end{tabular}
\end{table}

\begin{table}
  \caption{Arousal}
  \label{}
  \centering
  \begin{tabular}{lccc}
  \toprule
    Pretraining Strategy & Balanced Acc. & AUROC & F1 \\
    \toprule
    Contrastive CLIP-style (LOO) & \textbf{0.71 ± 0.002} & \textbf{0.776 ± 0.001} & 0.638 ± 0.002 \\ 
    Contrastive CLIP-style (Pairwise) & 0.708 ± 0.002 & 0.772 ± 0.001 & 0.627 ± 0.002 \\ 
    Contrastive SimCLR-style & 0.585 ± 0.048 & 0.616 ± 0.07 & 0.524 ± 0.027 \\ 
    MultiMAE & 0.604 ± 0.089 & 0.638 ± 0.136 & 0.613 ± 0.172 \\
    MultiMAE + Modality Drop & 0.637 ± 0.081 & 0.677 ± 0.128 & \textbf{0.641 ± 0.139} \\
    \bottomrule
  \end{tabular}
\end{table}


\end{document}